\documentclass{article}

\usepackage{hyperref}
\usepackage{graphicx} 
\graphicspath{{./figures/}}
\usepackage{amsmath} 
\usepackage{amssymb}  
\usepackage{bbm}
\usepackage{verbatim} 
\usepackage{algpseudocode}
\usepackage{tabulary}
\usepackage{algorithm} 
\usepackage[utf8]{inputenc} 
\usepackage{subcaption}
\usepackage{wrapfig}

\usepackage{bibunits} 
\usepackage{eso-pic}
\defaultbibliography{mybib} 

\usepackage[final,nonatbib]{corl_2018} 

\usepackage{color}

\newcommand{\algo}{Multi-DEX}
\newtheorem{definition}{Definition}

\title{Multi-objective Model-based Policy Search for Data-efficient Learning with Sparse Rewards}

\author{
  Rituraj Kaushik\\
  Inria, CNRS, Universit\'e de Lorraine\\
  \texttt{rituraj.kaushik@inria.fr}\\
  \And
  Konstantinos Chatzilygeroudis\\
  Inria, CNRS, Universit\'e de Lorraine\\
  \texttt{konstantinos.chatzilygeroudis@inria.fr}\\
  \AND
  Jean-Baptiste Mouret\\
  Inria, CNRS, Universit\'e de Lorraine\\
  \texttt{jean-baptiste.mouret@inria.fr}\\
}

\begin{document}
\AddToShipoutPicture*{\put(0,740){\parbox[b][\paperheight]{\paperwidth}{%
\vfill
\centering\footnotesize
Kaushik, R., Chatzilygeroudis, K. and Mouret, J.B. Multi-objective Model-based\\ Policy Search for Data-efficient Learning with Sparse Rewards. \\\textit{Conference on Robot Learning}, 2018, pp. 839-855.
}}}
\maketitle
\begin{bibunit}

\begin{abstract}
The most data-efficient algorithms for reinforcement learning in robotics are model-based policy search algorithms, which alternate between learning a dynamical model of the robot  and optimizing a policy to maximize the expected return given the model and its  uncertainties.
However, the current algorithms lack an effective exploration strategy to deal with sparse or misleading reward scenarios: if they do not experience any state with a positive reward during the initial random exploration, it is very unlikely to solve the problem.
Here, we propose a novel model-based policy search algorithm, \algo, that leverages a learned dynamical model to efficiently explore the task space and solve tasks with sparse rewards in a few episodes.
To achieve this, we frame the policy search problem as a multi-objective, model-based policy optimization problem with three objectives: 
(1) generate maximally novel state trajectories, 
(2) maximize the cumulative reward and 
(3) keep the system in state-space regions for which the model is as accurate as possible. We then optimize these objectives using a Pareto-based multi-objective optimization algorithm.
The experiments show that \algo{} is able to solve sparse reward scenarios (with a simulated robotic arm) in much lower interaction time than VIME, TRPO, GEP-PG, CMA-ES and Black-DROPS.

\end{abstract}



\section{Introduction}

Reinforcement Learning algorithms (RL) now allow artificial agents to learn to play many of the Atari 2600 games directly from pixels~\cite{mnih_human-level_2015} or to beat the world's best players at Go and chess with minimal human knowledge~\cite{silver2017mastering}. Unfortunately, these impressive successes rely on very long \emph{interaction times} between the algorithm and the system: for example, 38 days of play (real time) for Atari 2600 games~\cite{mnih_human-level_2015}, 4.8 million games for Go~\cite{silver2017mastering}, or about 100 hours of simulation time (much more for real time) for a 9-DOF mannequin that learns to walk~\cite{heess2017emergence}. This makes these algorithms well suited to simulated environments, but challenging to use with real robots, for which it is often not materially possible to perform more than few dozen of trials of a few seconds/minutes.

At the other end of the spectrum, model-based policy search algorithms can learn policies in a few minutes of interaction time, at the expense of a significant computation time between episodes. For instance, PILCO~\cite{deisenroth_gaussian_2015} and Black-DROPS~\cite{chatzilygeroudis2017black,chatzilygeroudis2018using} can learn a policy to balance a cart-pole or a policy to control a 5-DOF manipulator in less than 40-60 seconds of interaction time. However, these algorithms implicitly assume that most states can be associated to a positive or negative reward, as this is the case when the objective is to balance a pole or to follow a trajectory with a manipulator. In other words, these algorithms
are essentially greedy and mostly exploiting.

\begin{figure}
\centering
\includegraphics[width=0.85\linewidth]{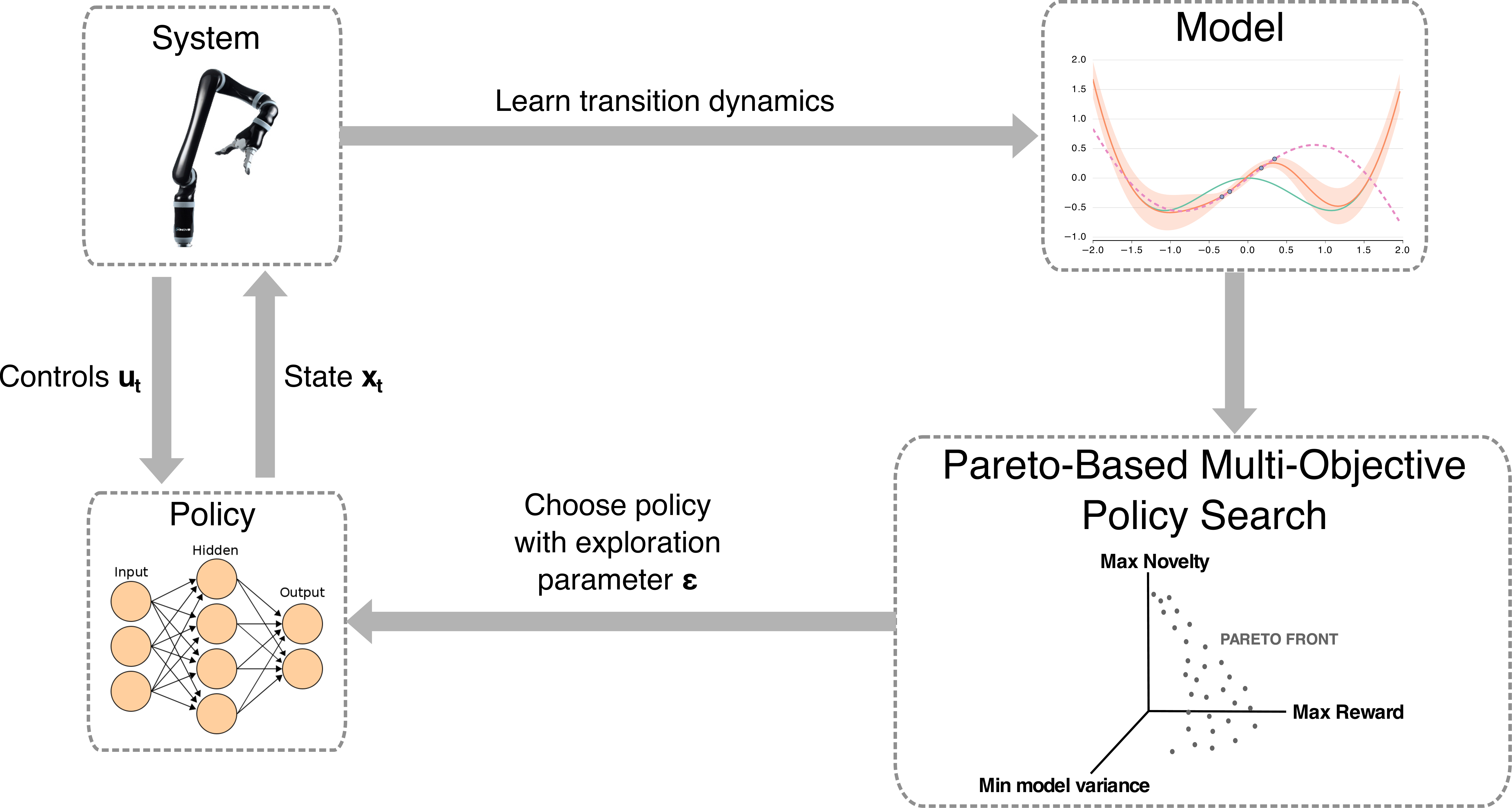}
\caption{\label{fig:concept} Overview of the \algo{} algorithm: The robot interacts with the environment with a policy in an episodic scheme; after each episode, \algo{} learns a GP model of the dynamics of the system and then performs multi-objective policy search to obtain a set of Pareto-optimal policies based on novelty, cumulative reward, and model variance; last, a policy is selected from the Pareto set based on an exploration parameter $\varepsilon$ and executed on the robot to collect new data.}
\vspace{-2em}
\end{figure}

While this assumption often holds in control tasks, rewards are much more sparse in many interesting learning scenarios: typically, we would like to reward the robot when it successfully achieves the task, and not for all the intermediate steps that we think should lead to success. For example, a robot might need to open a drawer and get rewarded by how much the drawer is open: in most of the state space, the reward is zero because the robot does not even touch the handle, meaning that PILCO or Black-DROPS need to open the drawer purely by chance to start learning a policy. In most realistic scenarios, this is unlikely to happen.

When there is no reward available, the robot should be ``curious'' and search for ``interesting stepping stones''. This question is central in reinforcement learning~\cite{kaelbling1996reinforcement, bellemare2016unifying,lopes2012exploration, andrychowicz2017hindsight,florensa2017reverse}, developmental robotics~\cite{oudeyer2007intrinsic,gottlieb2013information}, and evolutionary robotics~\cite{lehman2011abandoning,mouret_illuminating_2015,mouret_novelty-based_2011,doncieux2014beyond}. Among the most representative ideas, Novelty Search~\cite{lehman2011abandoning,mouret_novelty-based_2011} attempts to find behaviors that are as different as possible from what has been tested before; Self-Adaptive Goal Generation Robust Intelligent Adaptive Curiosity (SAGG-RIAC) selects random target states and searches for policies/trajectories to reach them; or Adaptive Curiosity pushes the robot towards situations in
which it maximizes its learning progress~\cite{oudeyer2007intrinsic,oudeyer2005playground}. Unfortunately, all these approaches require thousands (often more) learning episodes; in addition, they often focus solely on the exploration and often disregard that the robot might have a task to fullfil (and not only explore its body and the environment).

Our main insight is that we can leverage ideas from Novelty Search and developmental robotics but exploit them \emph{in a learned dynamical model} to keep the interaction time as low as possible. In essence, we would like to fuse the ``novelty-driven'' approaches with model-based policy search. At their core, modern model-based policy search algorithms alternate between learning a probabilistic dynamical model and optimizing a policy according to the model (and the uncertainty of the model)~\cite{deisenroth_gaussian_2015,chatzilygeroudis2017black}: In this paper, we propose to extend this idea to not only optimize for policies that are likely to work on the robot, but also for those that are predicted to be good for exploration. Our second insight is that we can optimize for novelty and cumulative reward using a Pareto-based Multi-Objective Evolutionary Algorithm (MOEA)~\cite{Deb2001MOU}, which simultaneously searches for all the Pareto-optimal trade-offs (instead of a single optimal solution when using an aggregated objective, \textit{Suppl. Mat.} Sec 2).

More concretely, we propose a novel model-based policy search algorithm, \algo{} (Multi-objective Data-Efficient eXploration), that efficiently explores the task space by framing the policy search problem as a multi-objective optimization problem with three objectives (Fig.~\ref{fig:concept}): (1) generate maximally novel state trajectories (2) maximize the cumulative reward and (3) keep the system in state-space regions for which the model is as accurate as possible. This optimization is performed with a Pareto-based multi-objective algorithm~\cite{Deb2001MOU} after each episode, once the dynamical model is updated with the new data. Thanks to the learned probabilistic dynamical model, we show that our proposed approach is highly data-efficient and typically needs only a few minutes of interaction time with the actual system to solve tasks with rare state transitions and sparse rewards. We compare the performance of \algo{} with recent algorithms for sparse reward scenarios (VIME~\cite{houthooft2016vime} and GEP-PG~\cite{Colas2018GEPPGDE}), a state-of-the-art model-free policy search algorithm (TRPO~\cite{schulman2015trust}), a state-of-the-art black-box optimizer (CMA-ES~\cite{hansen2006cma}), and a model-based policy search algorithm (Black-DROPS~\cite{chatzilygeroudis2017black}) in a sequential goal reaching task and in a drawer opening task; we show that \algo{} solves both tasks in order of magnitude less interaction time than the others.

\section{Problem Formulation}

\noindent We consider the following dynamical systems of the form:
\begin{align}
 \mathbf{x}_{t+1} = \mathbf{x}_t + f(\mathbf{x}_t,\mathbf{u}_t) + \mathbf{w}
\end{align}
with continuous-valued states $\mathbf{x}\in\mathbb{R}^E$ and controls $\mathbf{u}\in\mathbb{R}^F$, i.i.d. Gaussian system noise $\mathbf{w}$, and unknown transition dynamics $f$.
Our goal is to find a \emph{policy} $\pi(\mathbf{u}|\mathbf{x},\boldsymbol{\theta})$ parameterized by $\boldsymbol{\theta}\in\mathbb{R}^{\Theta}$ that maximizes the \emph{expected return} when following the policy $\pi_{\boldsymbol{\theta}}$ for $T$ time steps:
\begin{align}
  \label{eq:reward_j}
  J_r(\boldsymbol{\theta}) = \mathbb{E} \Bigg[\sum_{t=1}^{T}r(\mathbf{x}_t,\mathbf{u}_t,\mathbf{x}_{t+1}) \Big| \boldsymbol{\theta} \Bigg]
\end{align}
where $r(\mathbf{x}_t,\mathbf{u}_t,\mathbf{x}_{t+1})\in\mathbb{R}$ is the reward for being in state $\mathbf{x}_t$, taking action $\mathbf{u}_t$, and reaching state $\mathbf{x}_{t+1}$. We assume that $r$ can be sparse or may have plateaus; \emph{i.e.}, it can be zero for most values of $(\mathbf{x}_t,\mathbf{u}_t,\mathbf{x}_{t+1})$ or have large regions with constant value.

\section{Related Work}

\subsection{Encouraging Exploration in RL and Other Fields}

Exploration in trial-and-error learning has been investigated almost independently in several fields, with different inspirations and metaphors: in ``traditional'' reinforcement learning~\cite{kaelbling1996reinforcement,kearns2002near}, in deep reinforcement learning~\cite{andrychowicz2017hindsight,riedmiller2018learning,florensa2017reverse,Colas2018GEPPGDE}, in developmental robotics~\cite{forestier2016curiosity,forestier2017intrinsically}, and in evolutionary robotics~\cite{lehman2011abandoning,cully2018quality}.

In RL with discrete state-action spaces, Bayesian RL~\cite{kaelbling1996reinforcement} and PAC-MDP methods~\cite{kearns2002near} offer exploration with formal guarantees. When the state/action space is continuous, as this is usually the case in robotics, most successful approaches encourage exploration by performing \emph{action perturbation}~\cite{schulman2015trust} or on policy \emph{parameter perturbation}~\cite{hansen2006cma}. \emph{Directed exploration} approaches try to cover the whole state-action space as uniformly as possible. In this context, approaches such as~\cite{bellemare2016unifying,lopes2012exploration} try to drive the agent towards novel states. Hindsight Experience Replay (HER)~\cite{andrychowicz2017hindsight} exploits the fact that when experiencing a state-action trajectory, the goal being pursued influences the agent's actions but not the dynamics and therefore trajectories can be replayed with an arbitrary goal in an off-policy RL algorithm. Recently,~\cite{florensa2017reverse} proposed a ``reverse learning'' approach where the agent starts learning from a state nearby to the goal and gradually learns from more difficult initial states.  

In order to explore large and continuous state-action spaces efficiently, many approaches such as~\cite{pathak2017curiosity,houthooft2016vime,Schmidhuber1991CuriosityBoredom,mohamed2015variational} give the agent an additional reward called intrinsic reward that encourages the agent to perform actions that reduce the uncertainty in the agent's prediction ability. In this direction, VIME (Variational information Maximizing Exploration)~\cite{houthooft2016vime} was able to outperform (faster convergence and better solutions) many state-of-the-art RL algorithms such as TRPO~\cite{schulman2015trust} and ERWR~\cite{kober2011power} that use more basic exploration strategies.

In developmental and evolutionary methods, approaches such as novelty search~\cite{lehman2011abandoning}, quality-diversity~\cite{cully2018quality, mouret_illuminating_2015} and curiosity-driven learning~\cite{forestier2016curiosity} aim at uniformly covering an user-defined space called ``outcome space'', ``goal space'', or ``behavior space''. Approaches based on novelty-search encourage the agent to actively search for policies that produce novel outcomes. In this direction,~\cite{mouret_novelty-based_2011} has shown that by framing novelty and quality improvement of the policies as a multi-objective Pareto optimization problem, it is possible to evolve high quality policies compared to single objective counterparts. In curiosity-driven learning, the agent tries to accomplish goals progressively, based on increasing complexity. In this regime, 
IMGEP~\cite{forestier2017intrinsically} tries to discover policies that produce diverse outcomes by stochastically sampling different goals. Taking inspiration from IMGEP, GEP-PG~\cite{Colas2018GEPPGDE} fills the replay buffer of a DDPG algorithm~\cite{lillicrap2017continuous} with the data from initial exploratory episodes and then switches to a normal DDPG run.

Unfortunately, all the above-mentioned approaches need thousands of trials to find a working policy. For example, VIME needs more than trials $10000$ trials for the cart-pole swing-up task with sparse reward or HER~\cite{andrychowicz2017hindsight} needs around $80000$ trials to find a policy such that a manipulator pushes an object to a desired location. This makes them unsuitable for learning on real robotic systems. A few model-based approaches tried to address this problem by using an optimism-driven exploration in a model predictive control scheme~\cite{moldovan2015optimism,xie2016model}. Nevertheless, they are either too computationally intensive for online control~\cite{moldovan2015optimism} or require prior knowledge (\emph{e.g.}, dynamics equations) of the system~\cite{xie2016model}.

So far, to the best of out knowledge, no RL algorithm can efficiently (\emph{i.e.}, in little interaction time) explore in \emph{sparse-reward scenarios without the need of prior knowledge about the system morphology and its dynamics}.

\subsection{Model-based Policy Search}

The most data-efficient algorithms for RL in robotics are model-based policy search algorithms (MB-PS), which alternate between learning a dynamical model of the robot and optimizing a policy to maximize the expected return given the model and its uncertainties~\cite{chatzilygeroudis2017black,deisenroth_gaussian_2015,deisenroth_survey_2013}. Among the few proposed approaches, PILCO~\cite{deisenroth_gaussian_2015} and Black-DROPS~\cite{chatzilygeroudis2017black} exploit a probabilistic dynamical model to achieve high data-efficiency with minimal assumptions about the underlying system. For example, both algorithms require around 20 seconds of interaction time to solve the cart-pole swing-up task (at least 1-2 order of magnitude less than model-free policy search algorithms~\cite{deisenroth_survey_2013}).

PILCO uses gradient-based optimization by analytically computing the gradients of the reward with respect to the policy parameters and the probabilistic model (by using moment matching). Black-DROPS exploits the parallelization and the \emph{implicit averaging} properties of population-, rank-based optimizers (like CMA-ES~\cite{hansen2006cma}) to efficiently perform sampling-based evaluation of the trajectories in the model while accounting for the modelling uncertainties. Both algorithms are conceptually close but Black-DROPS is computationaly faster when several cores are available and does not impose any constraint on the type of the policy and reward functions~\cite{chatzilygeroudis2017black}.

Nevertheless, PILCO and Black-DROPS do not have an explicit exploration strategy besides a few random initial policies; they essentially \emph{exploit} whatever knowledge is captured in the model so far. As a result, when the reward function is sparse or deceptive~\cite{goldberg1987simple}, these algorithms struggle to find working policies or converge to sub-optimal ones: if high-reward states are associated with one or more \emph{rare state transitions} that are not observed \emph{by chance} in previous episodes, then these algorithms will fail to find the optimal policy because the rare transitions are not part of the model.

The main idea of the present paper is to combine the data-efficiency of MB-PS approaches with the creativity offered by novelty-based approaches to propose \emph{an algorithm that requires as little interaction time as possible even in scenarios where the reward is sparse or deceptive}.

\section{Approach}

\subsection{Learning system dynamics and reward model}
Gaussian Processes (GPs) are widely used for learning system dynamics because of their generalization ability and their ability to compute uncertainty along with the predictions~\cite{deisenroth_survey_2013,chatzilygeroudis2017black,deisenroth_gaussian_2015}. A GP is an extension of multivariate Gaussian distribution to an infinite-dimension stochastic process for which any finite combination of dimensions will be a Gaussian distribution~\cite{rasmussen2006gaussian}.

We use tuples of states $\mathbf{x}_t$ and actions $\mathbf{u}_t$ (\emph{i.e.}, $\mathbf{\tilde{x}}_t = (\mathbf{x}_t,\mathbf{u}_t)\in\mathbb{R}^{E+F}$) as training inputs for learning the system dynamics. The difference between the current and the next state vector, $\mathbf{\Delta}_{\mathbf{x}_t} = \mathbf{x}_{t+1}-\mathbf{x}_t\in\mathbb{R}^E$, is used as corresponding training targets. Then, $E$ independent GPs are used to model each dimension of the difference vector $\mathbf{\Delta}_{\mathbf{x}_t}$. 
%
%
Let $D_{1:t} = \{f(\mathbf{\tilde{x}}_1),...,f(\mathbf{\tilde{x}}_t)\}$ is a set of observations, we can query the GP at a new input point $\mathbf{\tilde{x}}_*$: 
%
  $p(\hat{f}(\mathbf{\tilde{x}}_*)|D_{1:t},\mathbf{\tilde{x}}_*) = \mathcal{N}(\mu(\mathbf{\tilde{x}}_*),\sigma^{2}(\mathbf{\tilde{x}}_*))$.
%
The mean and variance predictions of this GP are computed using a kernel vector $\pmb{k} = k(D_{1:t},\mathbf{\tilde{x}}_*)$, and a kernel matrix $K$, with entries $K_{ij} = k(\mathbf{\tilde{x}}_i,\mathbf{\tilde{x}}_j)$:
\begin{align}
  \label{eq:gp_mean}
  &\mu(\mathbf{\tilde{x}}_*) = \pmb{k}^{T}K^{-1}D_{1:t}\\
  \label{eq:gp_variance}
  &\sigma^{2}(\mathbf{\tilde{x}}_*) = k(\mathbf{\tilde{x}}_*,\mathbf{\tilde{x}}_*)-\pmb{k}^{T}K^{-1}\pmb{k}
\end{align}
In this paper, we use the exponential kernel with automatic relevance determination~\cite{rasmussen2006gaussian} and its hyper-parameters as well as the data noise hyper-parameter of the GP are found through Maximum Likelihood Estimation~\cite{rasmussen2006gaussian}. We use the Limbo C++11 library for GP regression~\cite{cully2018limbo}. In many cases, the immediate reward function might be unknown to the algorithm (\emph{i.e.}, not directly computable from the state/action vector). In such situations, we use Random Forest regression~\cite{breiman2001random} to learn the immediate reward function.

\subsubsection{Learning system dynamics with sparse transitions}

When the system dynamics have some sparse transitions that strongly affect the cumulative reward, learning the model in a na\"ive way (\emph{i.e.}, using all the data points) can lead to suboptimal models and the policy search will struggle to find good policies. For example, in our sequential goal reaching task (Sec.~\ref{sec:seq_goal_exp}) the state includes a Boolean switch that indicates whether the arm passed through the first way-point or not; because only one data point will have transition from $false$ to $true$ in any episode if the arm passes though the switch, learning a model with the full data (that mostly have no points with the switch transition) will most probably ignore these transition data points because they are likely to be (rightfully) considered as outliers. In other words, rare transitions have little impact on the likelihood or the fit of the dynamical model (as we additionally optimize the data noise hyper-parameter of the GP), whereas they are critical to leverage the dynamical model to learn a policy. 

The intuition here is to have a balanced blend of ordinary trajectories and trajectories with rare transitions (leading to high reward) for model learning. Learning a dynamics model this way will retain information not only about the rare and high rewarding transitions but also about the regions where no reward was observed. As a result, this model can efficiently be used for exploration as well as exploitation. To do this, we maintain two fixed sized experience buffers, $\mathbb{P}$ and $\mathbb{H}$, where we keep the trajectory data of the episodes for model learning. For each new trajectory executed on the robot, we insert a new observed trajectory into $\mathbb{P}$ in a FIFO fashion if no reward is observed; otherwise, we insert it into $\mathbb{H}$. Note that, we keep the maximum buffer size of $\mathbb{P}$ equal to the data size of $\mathbb{H}$ to have a uniform blend of ``high rewarding'' and ``no/low rewarding'' trajectories for model learning.

\subsection{Exploration-Exploitation Objectives}

The intuition behind \algo{} is to combine the data-efficiency of model-based policy search with the ability of diversity in novelty-based search algorithms to encourage exploration in policy search in a data-efficient way. Thus, we frame exploration as an additional objective that promotes policies that will most likely produce novel state trajectories in the real system. To keep the system within the more certain regions of the dynamics model so that prediction error can be avoided as much as possible, we set an additional objective to reduce the prediction variance in the trajectory. The three objectives are typically antagonistic, as the uncertainty is likely to be higher for more novel trajectories, and more novel trajectories are likely to not have an cumulative reward as high as those with the highest cumulative rewards (\textit{Suppl. Mat.} Sec 3).

\textbf{Cumulative Reward: } To compute the cumulative reward for a policy, we propagate through the learned GP model (with the mean prediction of the GPs, $f_{\mu}$) of the system for $T$ time steps using the policy, and observe the immediate reward in each step. Then, we aggregate all the immediate rewards to compute the cumulative reward (given the dynamical model) for the episode:

\begin{align}
  \label{eq:cum_rew}
  &\hat{J}_r(\boldsymbol\theta) = \sum_{t=1}^{T}r(\mathbf{x}_{t-1},\mathbf{u}_{t-1},\mathbf{x}_{t-1} + f_\mu(\mathbf{x}_{t-1}, \mathbf{u}_{t-1}))\\
  &\text{where }\mathbf{u}_{t-1} = \pi(\mathbf{x}_{t-1}|\boldsymbol{\theta})\nonumber
\end{align}
As it is evaluated in the model, optimizing this objective means exploiting the learned dynamics to find a policy that is likely to improve the cumulative reward.

\textbf{Novelty: } We compute the novelty score of a policy by comparing its expected state trajectory with the expected state trajectories of already executed policies (\textit{Suppl. Mat.} Sec 5). To keep the computation time tractable, we sub-sample them into $s_n$ equally spaced time-steps. We concatenate these state samples into one vector that we call \emph{state trajectory vector}, $\boldsymbol\beta$ which represents the ``expected state trajectory'' of a policy. These vectors are archived in a set $\mathbb{B}$ of fixed size $b_n$ and we re-compute them every time the GP model is updated. When $\mathbb{B}$ is full, we drop the least novel policy. We define the novelty score for any policy $\pi_{\boldsymbol\theta}$ as the minimum Euclidean distance of $\boldsymbol{\beta_\theta}$ to the state trajectory vectors in $\mathbb{B}$:

\begin{align}
  \label{eq:nov_score}
  \hat{J}_n(\boldsymbol\theta) = min(||\boldsymbol{\beta_\theta} - \boldsymbol\beta||^2)_{\forall \boldsymbol\beta \in \mathbb{B}}
\end{align}

This objective is promoting exploration policies that are likely to lead to state trajectories that are different from those already observed.

\textbf{Cumulative model-variance: } We define the cumulative model-variance for a policy $\pi_{\boldsymbol\theta}$ as the negative mean of the step-by-step model prediction variances:

\begin{align}
  \label{eq:var_score}
  &\hat{J}_{\sigma^2}(\boldsymbol\theta) = -\frac{1}{T}\sum_{t=1}^{T} (||\sigma_{\boldsymbol{x_t}}||^2)\\
  &\text{where }\boldsymbol{x_t}\text{ is given by applying the policy }\pi_{\boldsymbol{\theta}}\text{ on the model}\nonumber
\end{align}

This objective tries to keep the system close to regions where the model uncertainty is low and thereby avoids prediction error on the real system.

\subsection{Multi-Objective Optimization}

\begin{wrapfigure}[17]{R}{0.5\textwidth}
  \begin{minipage}{0.5\textwidth}
  \vspace{-2.5em}
  \begin{algorithm}[H]
    \scriptsize
    \caption{\algo}
    \label{algo:mops}
    \begin{algorithmic}[1]
      \Procedure{\algo}{}
        \State Define policy $\pi: \mathbf{x}\times\boldsymbol{\theta}\to\mathbf{u}$
        \State Define $\varepsilon \in [0,1]$ \Comment{\emph{User defined exploration parameter}}
        \State $D = \emptyset$
        \For {$i=1\to N_R$} \Comment{\emph{$N_R$ random episodes}}
          \State $\boldsymbol{\theta_{rand}}$ = random\_policy()
          \State Execute $\pi_{\boldsymbol{\theta_{rand}}}$ on the real system
          \State Collect $(\mathbf{x}_t,\mathbf{u}_t)\to\mathbf{x}_{t+1}$ and update $\mathbb{P}$, $\mathbb{H}$ and $D$
        \EndFor
  
        \While{$task \ne solved$} \Comment{\emph{Learning episodes}}
          \State Train dynamics model with GPs using collected data $D$
          \State Update $\mathbb{B}$ using the trained GPs
          \State Get Pareto set of policies $\pi_{\boldsymbol{\theta}}$ using NSGA-II according to $\hat{J}_r, \hat{J}_n, \hat{J}_{\sigma^2}$
          \State $\pi_{\boldsymbol{\theta}^*} = $ policy in Pareto front that has maximum $\hat{J}_r$
          \State With probability $\varepsilon$, replace $\pi_{\boldsymbol{\theta}^*}$ by randomly selecting from the Pareto front
          \State Execute $\pi_{\boldsymbol{\theta}^*}$ on the real system
          \State Collect $(\mathbf{x}_t,\mathbf{u}_t)\to\mathbf{x}_{t+1}$ and update $\mathbb{P}$, $\mathbb{H}$ and $D$
        \EndWhile
  
      \EndProcedure
    \end{algorithmic}
  \end{algorithm}
  \end{minipage}
  \end{wrapfigure}

Most algorithms aggregate objectives using a weighted sum. However, this means that the weights have to be chosen beforehand, which requires a time-consuming trial-and-error process; moreover, some optimal trade-offs cannot be found using a weighted sum~\cite{Deb2001MOU}. Here, we rely on the multi-objective optimization literature~\cite{Deb2001MOU} and the Pareto-optimality concept:

\vspace{-0.6em}
\begin{definition}
  A solution $x_{1}$ is said to dominate another solution $x_{2}$, if:
  \vspace{-0.75em}
  \begin{enumerate}
    \item the solution $x_{1}$ is not worse than $x_{2}$ with respect to all objectives,
    \vspace{-0.5em}
    \item the solution $x_{1}$ is strictly better than $x_{2}$ at least one objective.
  \end{enumerate}
\end{definition}
Instead of searching for a single, optimal policy, a multi-objective optimization algorithm searches for the set of non-dominated policies in the search space, called the Pareto set: these policies are Pareto-optimal trade-offs between the objectives. For instance, a novel policy $\pi_1$ with a low cumulative return is as interesting as less novel policy $\pi_2$ but with a better cumulative return; however, a policy $\pi_3$ which is both less novel than $\pi_1$ and lower-performing is not interesting and is not Pareto-optimal.

While the set of non-dominated solutions is a very small subset of the search space, there are still many policies to choose from at the end of the optimization process. \algo{} executes on the robot the policy with the maximum reward \emph{from the Pareto set} with a probability $1 - \varepsilon$ (exploitation) and choose a random policy \emph{from the Pareto set} with a probability $\varepsilon$ (exploration). As a consequence, if two policies have the same cumulative return, then \algo{} will always choose the most novel and less uncertain one; and if two policies are equally novel, it will always choose the one with the best cumulative return and uncertainty; and among the Pareto-optimal policies, it will select either the one with the best cumulative return or a random policy from the Pareto set.
To find the Pareto set according to the dynamical model, we use the NSGA-II algorithm~\cite{deb2002fast} (using the Sferes2~\cite{mouret2010sferes} C++ library), which is a widely used, state-of-the-art algorithm (see \textit{Suppl. Mat.} Sec 1 for details about NSGA-II).

Overall, \algo{} starts with random policy and executes it on the robot (Algo.~\ref{algo:mops}). Then it enters an iterative loop, where \algo{} (1) learns a model of the dynamics of the system and a model of the reward function (if needed) from the data collected from previous episodes, (2) finds a set of non-dominated policies, using NSGA-II, that maximizes the cumulative reward (Eq.~\ref{eq:cum_rew}), novelty (Eq.~\ref{eq:nov_score}) and minimizes the model prediction variance (Eq.~\ref{eq:var_score}), and (3) selects the policy that corresponds to maximum reward (with a probability $\varepsilon$, it replaces it with a randomly selected policy from the Pareto front) and executes it on the robot. This iterative process continues until the task is solved (Algo.~\ref{algo:mops}).

\section{Experiments}

We compare to several state-of-the-art approaches in a sequential goal reaching task and in a drawer opening task: (1) Black-DROPS~\cite{chatzilygeroudis2017black}, a model-based policy search algorithm; (2) TRPO~\cite{schulman2015trust}, a model-free policy gradient approach; (3) TRPO with the VIME exploration strategy; (4) CMA-ES~\cite{hansen2006cma}, a black-box optimizer effective for direct policy search, and (5) GEP-PG, a curiosity-driven model-free approach. In all the tasks, we give a budget of 2500 trials to all the model-free approaches and 250 trials to the model-based ones (see \textit{Suppl. Mat.} Sec 6 and Sec 7 for further experimental details and additional experiments). 
The \algo{} source code and the supplementary video are available online.\footnote{\algo{} source code: \url{https://github.com/resibots/kaushik_2018_multi-dex}}\footnote{Video: \url{https://youtu.be/XOBWq7mkYho}}

\subsection{Sequential goal reaching with a robotic arm}
\label{sec:seq_goal_exp}

\begin{figure}
  \centering
  \begin{subfigure}{.35\textwidth}
    \centering
    \includegraphics[width=0.8\textwidth]{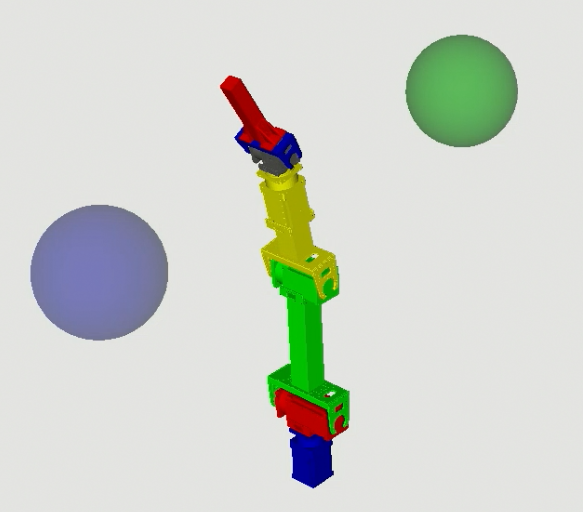}
    \caption{\label{fig:seq_goal_task}}
  \end{subfigure}%
  \hfill
  \begin{subfigure}{0.65\textwidth}
    \centering
    \includegraphics[width=0.9\textwidth]{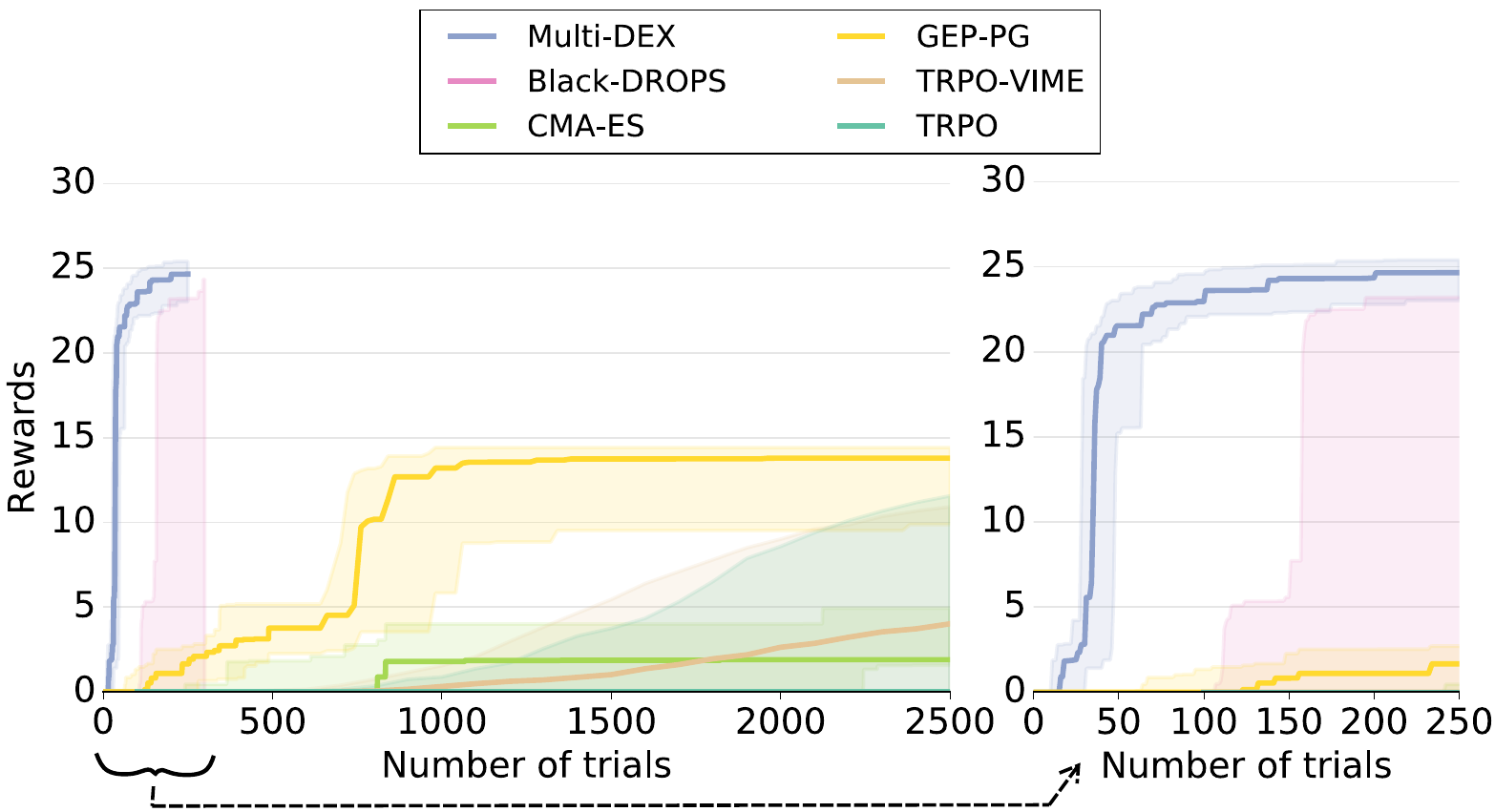}
    \caption{\label{plot:seq_goal_task}}
  \end{subfigure}
  \caption{(a) Setup of the sequential goal reaching task; the goal of the 2-DOF simulated arm is to reach with its end-effector the green goal while first passing through the blue region. (b) Best reward found per trial (20 replicates). The lines are median values and the shaded regions the 25th and 75th percentiles.
  \algo{} significantly outperforms all the other approaches and finds working policies in about only 7 minutes of interaction (around 100 episodes/trials).}
  \vspace{-1.25em}
\end{figure}

This simulated task consists of a robotic arm with 2 degrees of freedom (Fig.~\ref{fig:seq_goal_task}). The goal of the arm is to to reach the green goal with its end-effector while first passing through the blue region. The length of each episode is 4 seconds with a control frequency of $10$Hz. All the algorithms use feed-forward neural network policies. In more detail:

\begin{itemize}
  \setlength\itemsep{0.05em}
  \item \textbf{State:} $\mathbf{x}_{\text{seq}} = [\theta_{0}, \theta_{1}, \gamma] \in \mathbb{R}^3$ , where $\theta_{0}, \theta_{1}$ are joint angles and $\gamma$ is a boolean value which is set to 1 for rest of the episode if the end-effector has touched the blue region.
  \item \textbf{Actions:} $\mathbf{u}_{\text{seq}} = [v_{0}, v_{1}] \in \mathbb{R}^{2}, -2\le v_{0}, v_{1} \le 2\text{ }rad/s$ are joint velocity commands.
  \item \textbf{Reward:} The reward function depends only on the current state and is unknown to the algorithm as it is not a \emph{direct} function of observable state:
  \begin{align}
    r(\mathbf{x}_{\text{seq}}) =
                                \begin{cases}
                                  0, &\quad\text{if } \gamma \ne 1 \text{ or } \Delta \ge 0.1\\
                                  \exp \big (-\alpha\Delta^2 \big ), &\quad\text{otherwise.} \\
                                \end{cases}
    \end{align}
  where $\Delta$ is the Euclidean distance between green goal and the end-effector of the arm and $\alpha$ is a user-defined constant ($\frac{0.5}{0.2*0.2}$ in our experiments).
  After every episode, \algo{} learns the reward function from the collected data using Random Forest regression~\cite{breiman2001random}.
\end{itemize}
\algo{} is able to find high-performing policies in less than 100 trials on the target system (around 7 minutes of interaction), whereas none of the other approaches are able to achieve similar rewards even after 2500 trials (Fig.~\ref{plot:seq_goal_task} --- 20 replicates). As this task has a highly sparse reward space, CMA-ES, Black-DROPS and TRPO show almost zero median reward.

\subsection{Drawer opening task with a robotic arm}
\label{sec:drawer_open_exp}

\begin{figure}
  \centering
  \begin{subfigure}{.3\textwidth}
    \centering
    \vspace{2em}
    \includegraphics[width=0.8\textwidth]{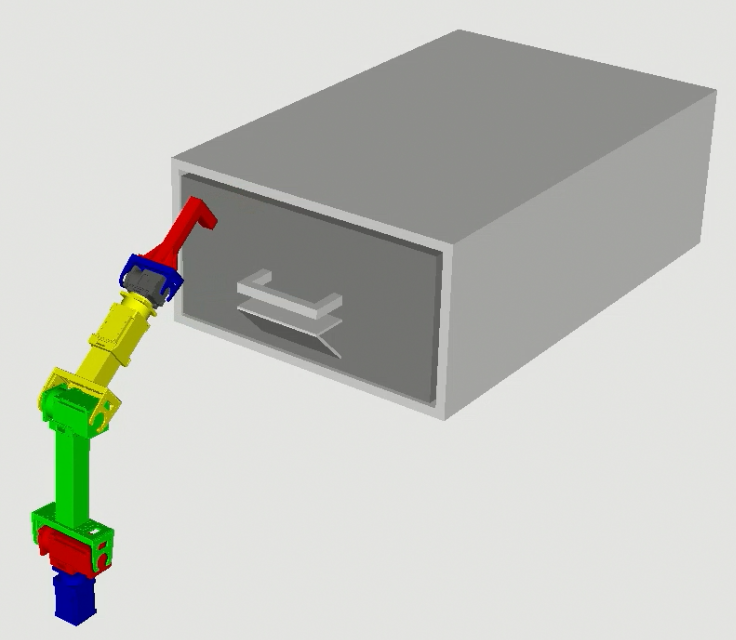}
    \caption{\label{fig:drawer_opening_task}}
  \end{subfigure}%
  \hfill
  \begin{subfigure}{0.7\textwidth}
    \centering
    \includegraphics[width=0.9\textwidth]{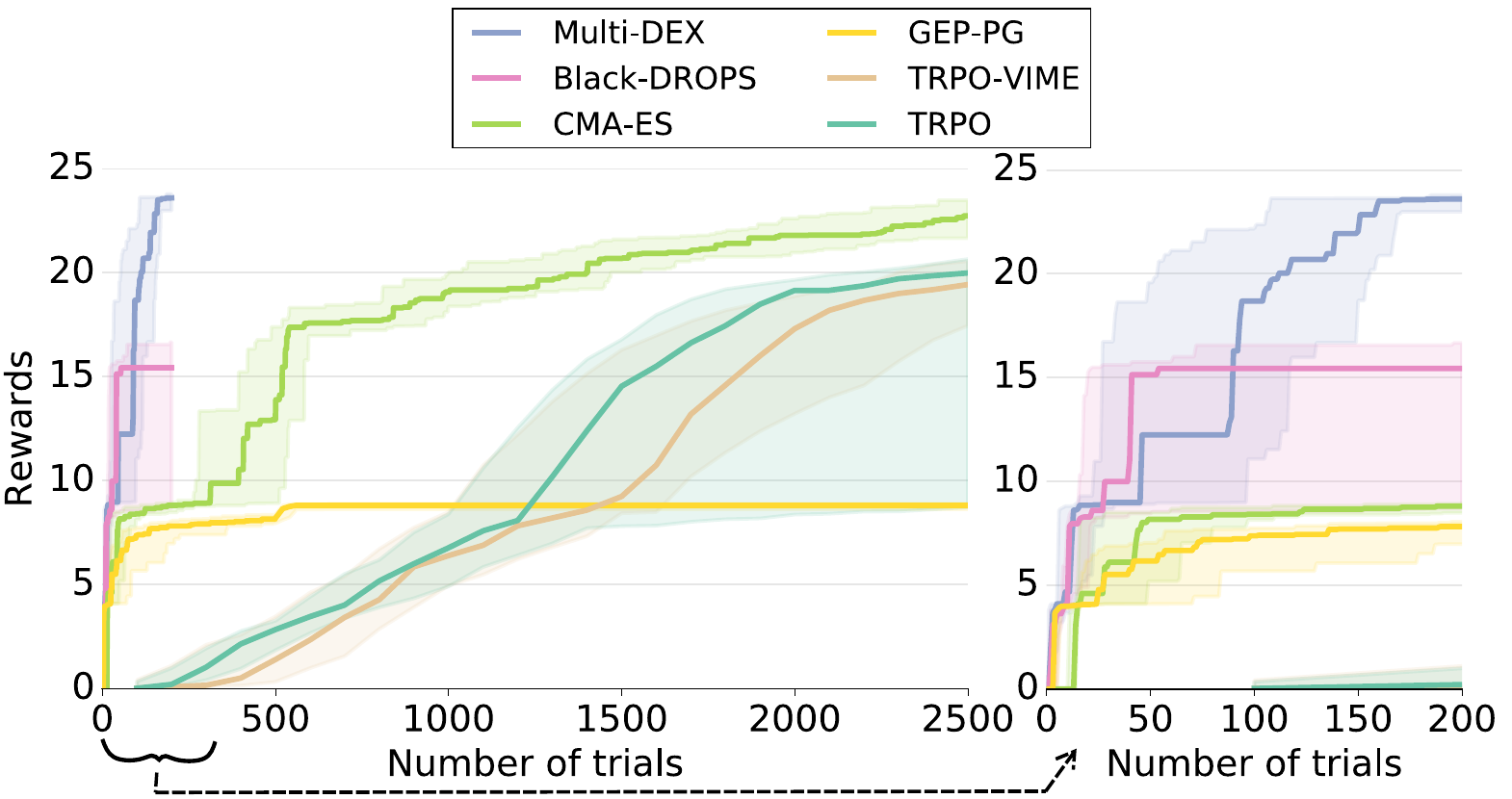}
    \caption{\label{plot:drawer_opening_task}}
  \end{subfigure}
  \caption{(a) Setup of the drawer opening task; the goal of the 2-DOF arm is to open the drawer and go back to the up-right position. (b) Best reward found per trial (20 replicates). The lines are median values and the shaded regions the 25th and 75th percentiles.
  \algo{} outperforms all the other approaches and finds working policies in about only 14 minutes of interaction (200 trials).}
  \vspace{-1.25em}
\end{figure}

The goal of this simulated task is to open a drawer with a 2-DOF robotic arm and go back to the up-right position (Fig.~\ref{fig:drawer_opening_task}). Similarly to the previous task, the length of each episode is 4 seconds ($10$Hz control) and all the algorithms use neural network policies. In more detail:

\begin{itemize}
  \setlength\itemsep{0.05em}
  \item \textbf{State:} $\mathbf{x}_{\text{drawer}} = [\theta_{0}, \theta_{1}, \delta] \in \mathbb{R}^3$ , where $\theta_{i}$ are joint angles and $\delta$ the drawer displacement.
  \item \textbf{Actions:} $\mathbf{u}_{\text{drawer}} = [v_{0}, v_{1}] \in \mathbb{R}^{2}, -1\le v_{0}, v_{1} \le 1\text{ }rad/s$ are joint velocity commands.
  \item \textbf{Reward:} In this task, the reward depends only on the current state and is known to the algorithm as it is a function of the observable state. The total reward is given by:
  \begin{align}
    & r(\mathbf{x}_{\text{drawer}}) = r_{\text{return}}(\mathbf{x}_{\text{drawer}}) + \delta\\
    & r_{\text{return}}(\mathbf{x}_{\text{drawer}}) =
                              \begin{cases}
                                0, &\quad\text{if } \delta \le 0.2\\
                                \exp (-\theta_{0}^{2} - \theta_{1}^{2}), &\quad\text{otherwise.} \\
                              \end{cases}
  \end{align}
\end{itemize}

In this task, the reward space is moderately sparse compared to the previous one; however, it has a misleading reward space because of the composition of two different rewards. Any algorithm without efficient exploration will converge to the reward associated with the drawer displacement only. \algo{} is able to find policies that complete the task in just 200 trials (around 14 minutes of interaction), whereas all the other approaches are deceived by the reward space (Fig.~\ref{plot:drawer_opening_task} --- 20 replicates). Black-DROPS and TRPO, without any directed exploration, fall in the sub-optimal solution (\emph{i.e.}, just opening the drawer quickly). CMA-ES, TRPO-VIME and GEP-PG either fail to converge or need more than 2500 trials to reach the same quality of solutions as \algo{}.

\section{Conclusion}
It is well established that learning a dynamical model can make policy search highly-data efficient for control tasks~\cite{chatzilygeroudis2017black,deisenroth_survey_2013} when (1) the state-space is low-dimensional enough to learn a model, (2) the model is probabilistic so that the uncertainty of the predictions can be taken into account, and (3) it is possible to rely on a large computation time between episode (for the policy optimization stage). In this paper, we showed that the same model can be leveraged to improve exploration for tasks in which the reward is sparse or deceptive: compared to state-of-the-art approaches to exploration for policy search, which are all model-free, our model-based policy search algorithm requires at least an order of magnitude fewer episodes and reach higher rewards in all our experiments.



\clearpage
\acknowledgments{This work received funding from the European Research Council (ERC) under the European Union’s Horizon 2020 research and innovation programme (GA no. 637972, project ``ResiBots''), and the European Commission through the project H2020 AnDy (GA no. 731540).\vspace{-1em}}



\putbib
\end{bibunit}
\clearpage

\setcounter{section}{0}
\begin{bibunit}
\begin{center}
\LARGE \textbf{Supplementary Material \\   }
\end{center}


\section{Non-dominant Sorting Genetic Algorithm - II (NSGA II)}
We use NSGA-II for multi-objective policy optimization.

NSGA-II starts with random initialization of parent population $P_0$ of size $N$. $P_0$ is sorted based on non-domination and a fitness score (or rank) is assigned to each of the individuals. For example, solutions which are not dominated by any solutions in all the objectives are assigned a fitness or rank 0 (i.e $0^{th}$ level of domination). Similarly, solutions which are dominated in all the objectives by rank 0 solutions are assigned a fitness value or rank of 1 (i.e $1^{st}$ level of domination) and so on. Then $N/2$ individuals are selected with binary tournament selection based on the fitness/rank scores (lower the better) and then applied binary crossover and mutation operators to obtain a offspring population $Q_0$ of size $N$. For each generation $k$, the algorithm performs the following steps:

\begin{enumerate}
\item A combined population $R_{k} = P_{k} \cup Q_{k}$ of size $2N$ is formed.

\item $R_k$ is sorted in ascending order based on the level of non-domination into sets $F_0$, $F_1$, $F_2$ and so on. Then solutions are selected starting from $F_0$ to the last possible level (say $F_l$) to form a new parent population $P_{k+1}$ of size $N$. If last set $F_l$ cannot be accommodated completely in $P_{k+1}$, then the solutions from $F_l$ is selected based on crowded comparison operator $\prec_n$ ~\cite{deb2002fast}.

\item Selection, crossover (sbx) and polynomial mutation operators are applied on the new parent population $P_{k+1}$ to produce the new offspring population $Q_{k+1}$ of size $N$.

\item The algorithm continues until a specified number of generation is reached. Then it returns the set of non-dominated solutions (approximation of the true Pareto-front).
\end{enumerate}

\section{Why Pareto optimality should be preferred over weighted sum objectives?}
When optimizing several objectives with a weighted sum, the weights determine the trade-off between the objectives (e.g., more exploration vs more exploitation) and the optimum is a particular Pareto-optimal solution. The important question for multi-objective optimisation is therefore: can any Pareto-optimal solution (i.e., any optimal trade-off) be found by selecting the right combination of weights? The answer depends on the shape of the Pareto front (the set of Pareto-optimal solutions). If it is convex, then all the Pareto-optimal solutions can be found by varying the weights; however, if it is not, then some Pareto-optimal trade-offs cannot be found using a weighted sum ~\cite{Deb2001MOU}. In the learning problems tackled in this paper, there is no formal guarantee that the Pareto front is convex, therefore it is theoretically better to use another approach to find a suitable Pareto-optimal trade-off. In other words, the best trade-off between reward, novelty, and model variance might not be expressible with a combination of weights. In addition, setting weights requires a good normalization method (since the trade-off will be affected by the scale of each objective), whereas a Pareto-based approach will usually be independent of the scale.

For completeness, we compared experimentally our Pareto-based approach with both fixed and stochastic weighted aggregation of the objectives to a single objective. For fixed weighted aggregation we used weights 0.4, 0.3 and 0.3 for novelty, variance and cumulative reward respectively. On the other hand, for stochastic weighted aggregation we used 4 sets of weights: $\{0.6, 0.3, 0.1\}$, $\{0.1, 0.3, 0.6\}$, $\{0.1, 0.45, 0.45\}$ where each element of the sets corresponds to the weights for novelty, variance and cumulative reward respectively. We randomly select a set of these weights for each episode and use it for aggregation of the objectives. The results show that weighted aggregation doesn't give consistent results on both the experiments. Although a different set of weights could give different results, in the drawer opening task Pareto-based approach out performs both the variants of weighted aggregation approach with higher mean performance and smaller variance over the replicates. In the sequential goal reaching task the stochastic weighted aggregation shows similar performance to the Pareto-based approach. However, fixed weighted aggregation approach did not solve the task at all.

\begin{figure}
  \centering
  \begin{subfigure}{.52\textwidth}
    \centering
    \includegraphics[width=0.8\textwidth]{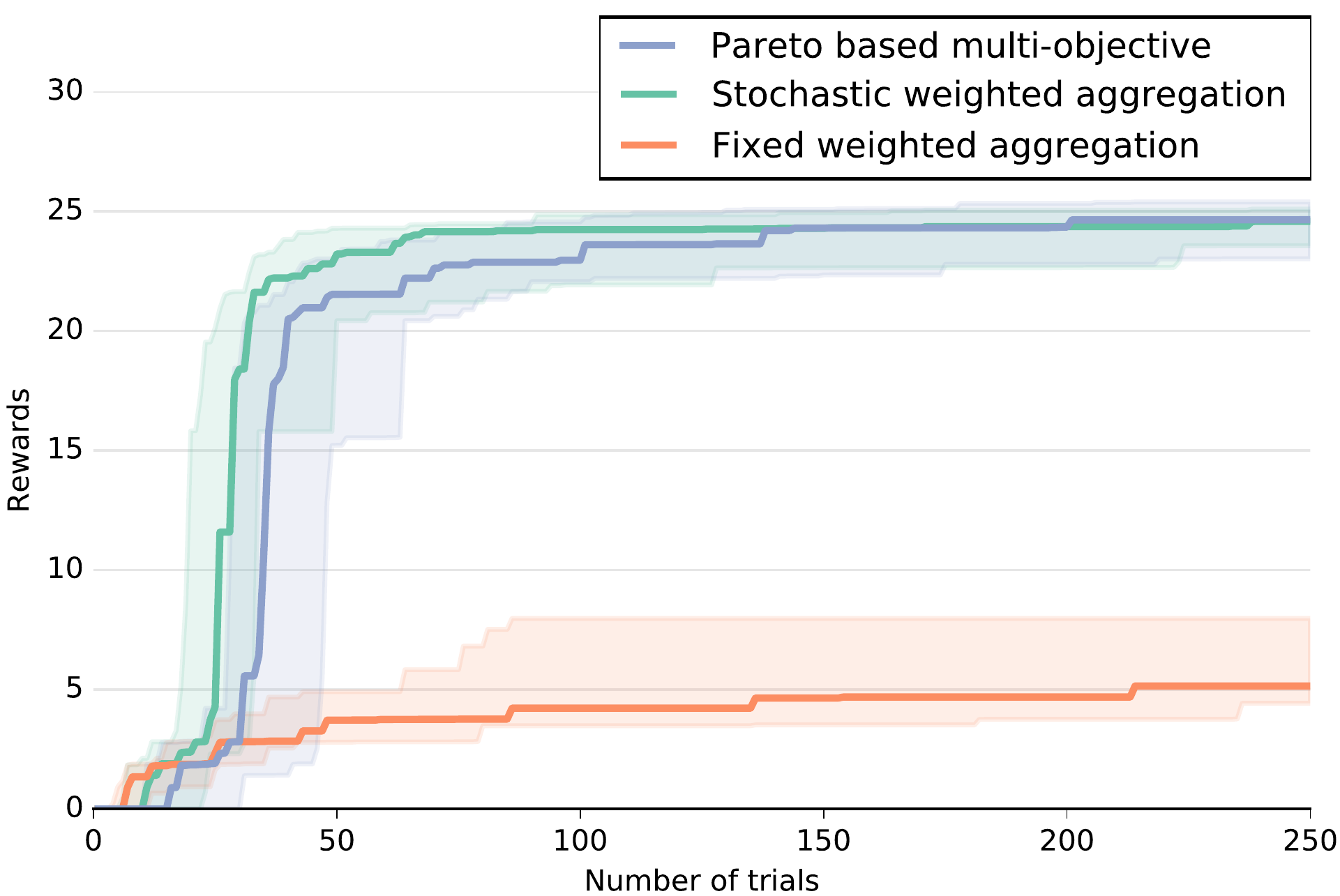}
  \caption{\label{plot:Sequential goal reaching task weighted sum vs pareto comparision} Sequential goal reaching task }
  \end{subfigure}%
  \hfill
  \begin{subfigure}{0.48\textwidth}
    \centering
    \includegraphics[width=0.8\textwidth]{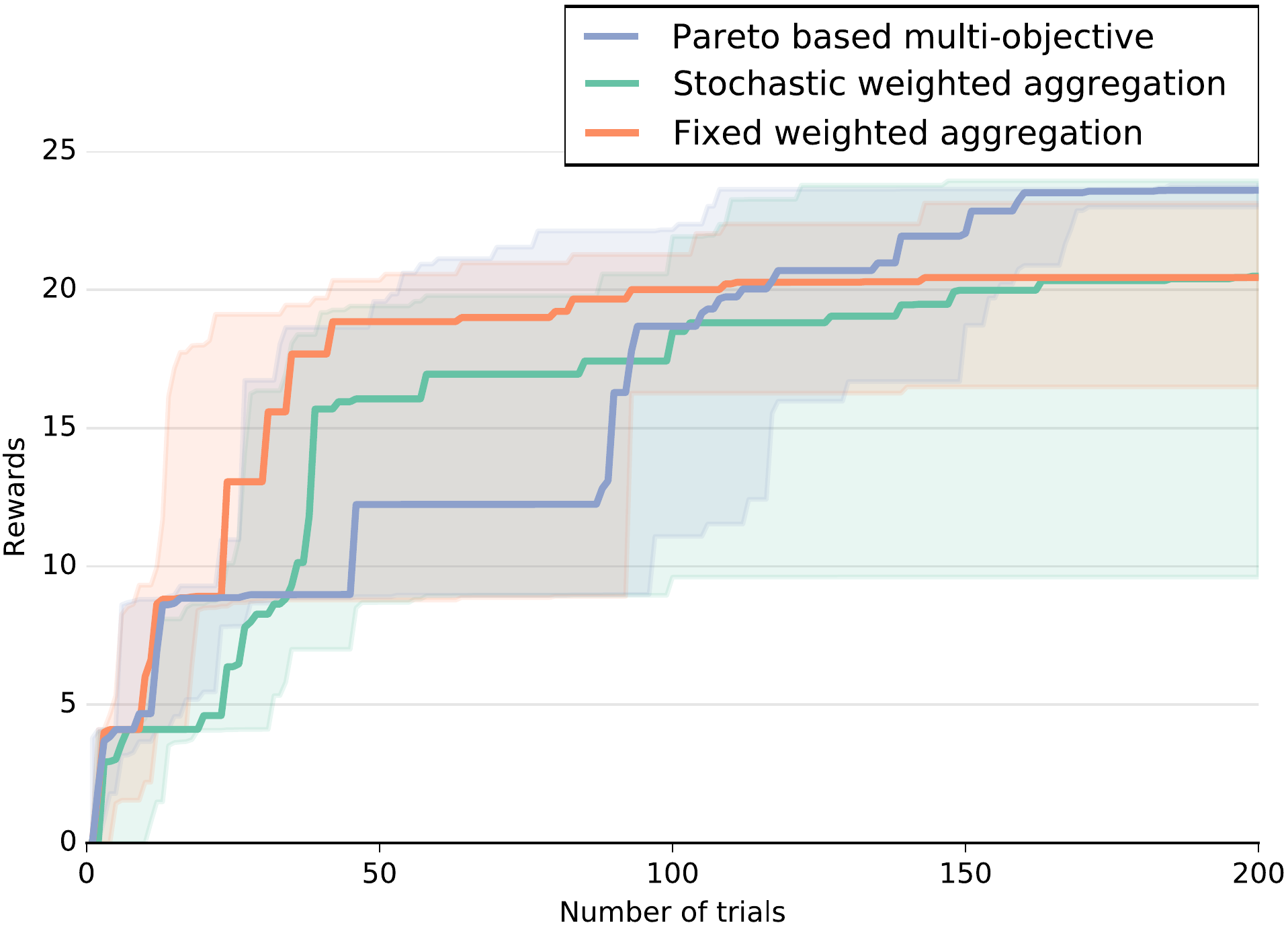}
    \caption{\label{plot:Drawer opening task weighted sum vs pareto comparision} Drawer opening task}
  \end{subfigure}
  \caption{Best reward found vs number of trials plots show the comparison between Pareto based multi-objective approach and weighted aggregation approach. For stochastic weighted aggregation approach we defined 4 set of weights with different trade-offs among the objectives. For each episode we randomly select one set of weights and use them for aggregation of the objectives into a single objective. On the other hand, for the fixed weighted aggregation approach we defined a single set of weights and used them for the aggregation of the objectives.}
  \vspace{-1.25em}
\end{figure}

\section{Interaction of the objectives among themselves}
The first objective in multi-DEX is to find trajectories (i.e., corresponding policies) that have high reward. The second objective tries to ensure that for a particular value of reward we get as novel trajectory as possible. Now these two objectives might be antagonistic and for most of cases both cannot be improved without compromising the other. Thus we want a Pareto-optimal set of policies that gives the trade-off between these two objective, starting from zero reward to maximum reward possible on the model. However, as we are using GP mean prediction only to find trajectories, it is possible that we find policies that drive the system though highly uncertain regions on the model. If this is the case then those policies will have completely different behavior on the system from the one predicted using the model. Thus we have the third objective that tries to keep the trajectories as close as possible to the more certain regions of the model. Again this new objective might be antagonistic to previous two objectives in some cases. Therefore, we want a Pareto-optimal trade-off among all the objectives. It is to be noted here that in Multi-DEX novelty score is highly dependent upon the order in which the states are visited and not that much on the individual states. So it is possible to have a high novelty score by going through similar states that were observed before (i.e., points with less uncertainty) but in a completely different order. Once the Pareto-optimal trade-offs are identified, all of them are potentially good candidates. This is why we alternatively take a random policy from these good candidate (minimize bias in exploration) or the one with the best predicted reward (exploitation).

\section{ How scalable is the multi-objective optimization method (NSGA-II) with regards to the number of parameters of the policy?}
Several recent papers show that evolutionary algorithms, such as NSGA-II, can be competitive to optimize deep neural network policies in reinforcement learning scenarios~\cite{petroski2017deep,salimans2017evolution}. Thus, we believe that NSGA-II should be able to scale up to higher-dimensional policy optimization, at the obvious expense of more optimization time. At any rate, it should be emphasized that what matters the most in this work is the interaction time with the system/robot, and not the policy optimization time that happens on the learned dynamical model.

\section{Why Expected Trajectory for Novelty Score?}
In \algo{} we have defined the novelty score of any policy as the minimum Euclidean distance between the expected state trajectory of that policy to all the \emph{expected state trajectories} of the executed policies on the system. There is a strong reason for keeping expected state trajectory vectors and not the observed state trajectory vectors on the system to compute the novelty score of a policy. Because of the imperfection in the model, the model might predict unachievable state trajectories to maximize novelty, which might give a completely different state trajectory on the actual system than the one expected by the model. Now, if we compute the novelty score for a policy by comparing with the observed state trajectory vector on the system, then the already tried policies predicting unachievable behaviors will have very high novelty scores. As a result, optimizer might return these trajectories again and again for every iteration of the algorithm.

A solution to this problem is to keep the expected state trajectory vectors instead to compute the novelty scores. As as result, the unachievable state trajectories that were already tried on the system will have low novelty scores; and hence, policies with these state trajectories will be rejected because of low novelty score by the optimizer.

\section{Details on the Experimental Setup}
\subsection{Simulator and source code}
For all the experiments we have used DART physics simulation library~\cite{lee2018dart} to implement the scenarios. For TRPO, VIME and GEP-PG we used python version of the DART library (pydart2\footnote{\url{https://github.com/sehoonha/pydart2}}); for CMA-ES, Black-DROPS and \algo{} we used the C++ version of the library for simulation.

For CMA-ES, we used Limbo's wrapper of libcmaes library\footnote{\url{https://github.com/beniz/libcmaes}}. For all other algorithms (GEP-PG, TRPO, TRPO-VIME and Black-DROPS), we used the code provided by the authors and integrated our own scenarios. The source code of \algo{} can be found here: \url{https://github.com/resibots/kaushik_2018_multi-dex}

\subsection{General and Exploration Parameters}
For GEP-PG, we used 500 exploratory episodes for the ``Drawer Opening'' and ``Sequential Goal Reaching'' tasks; and 100 exploratory episodes for ``Pendulum Swing-up'' task. For all the experiments with GEP-PG we used 10 initial bootstrap episodes.

For TRPO-VIME we used $\eta=0.001$ for the ``Drawer Opening'' task. For the ``Sequential Goal Reaching'' and ``Pendulum Swing-up'' tasks, we used $\eta=0.001$.

For CMA-ES, we used the active-IPOP version with elitism enabled~\cite{auger2005restart,loshchilov2013cma}. We used 3 restarts and $2500$ maximum function evaluations.

For \algo, we used $\varepsilon=0.3$ for the ``Sequential Goal Reaching'' and ``Pendulum Swing-up'' tasks. We used $\varepsilon=0.4$ for the ``Drawer Opening'' task. In all the experiments, to compute the novelty score, we sub-sample the expected state trajectories at 10 equally spaced points in time. In all the cases, we bootstrapped the algorithm with 5 initial random trials.

\subsection{Policy and Parameter Bounds}
For all the experiments, we used neural network policies. For Black-DROPS, GEP-PG and \algo{} in the ``Sequential Goal Reaching'' and ``Drawer Opening'' tasks, we used one hidden layer with $5$ neurons and in the ``Pendulum Swing-up'' task we used one hidden layer with $10$ neurons. For TRPO and TRPO-VIME we used two hidden layers with $10$ neurons each (the implementations that we used required at least 2 hidden layers for the neural network policies).

The policy parameter bounds used in \algo{} are $[-1,1]$ for ``Sequential Goal Reaching'', $[-3,3]$ for ``Drawer Opening'' tasks and $[-5,5]$ for ``Pendulum Swing-up'' task.

\subsection{NSGA-II Parameters}
We used the Sferes2 C++ library's~\cite{mouret2010sferes} implementation of NSGA-II algorithm in \algo. The parameters used for NSGA-II are the following:

\begin{itemize}
  \item crossover rate = 0.5
  \item mutation rate = 0.1
  \item $\eta_m$ = 15.0
  \item $\eta_c$ = 10.0
  \item mutation type: Polynomial~\cite{Deb2001MOU}
  \item cross-over type: Simulated Binary Crossover~\cite{deb1999self}
\end{itemize}

We run NSGA-II for $600$ generations with population sizes of $200$ for the ``Drawer Opening'' and ``Sequential Goal Reaching'' tasks and $300$ for the ``Pendulum Swing-up'' task.

For the first episode, the initial population of policies is randomly initialized. From the 2nd episode onwards, we insert some policies from the Pareto-front into the initial population. The maximum number of these inserted policies is not more than $30\%$ of the total population.

\subsection{Gaussian Process Model learning}
In this work, we used the Limbo C++ library's~\cite{cully2018limbo} implementation of Gaussian process regression for probabilistic dynamics model learning. We used exponential kernel with automatic relevance determination~\cite{rasmussen2006gaussian}. We optimize the hyper-parameters of the kernel via Maximum Likelihood Estimation using the Rprop optimizer~\cite{riedmiller1992rprop,blum2013optimization}.

We maintain two experience buffers (of fixed size) $\mathbb{P}$ and $\mathbb{H}$ to keep the state-action trajectories for model learning (see the main text).

\section{Additional Experiments}
\subsection{Deceptive pendulum swing-up task}

This simulated task consists of a pendulum powered by an underpowered torque-controlled actuator. The goal in this task is to swing the pendulum to the upright position applying torques as small as possible (\emph{i.e.}, using minimum power) to the actuator and hold it in that position. The learning algorithm gets a constant positive reward of +10 every time-step the pendulum is in upright position (within some region). It also gets a negative reward proportional to the square of torque for every time step. Because of this two rewards, the total reward function possesses a deceptive landscape and algorithms without efficient exploration will converge to a reward of zero which is achieved when no torque is applied to the pendulum. This type of \emph{``Gradient Cliff''} reward landscape was first proposed by~\cite{lehman2017more}. The control frequency in this task is $10Hz$ and every episode is 4 seconds long. In more detail:

\begin{figure}
  \centering
  \includegraphics[width=\textwidth]{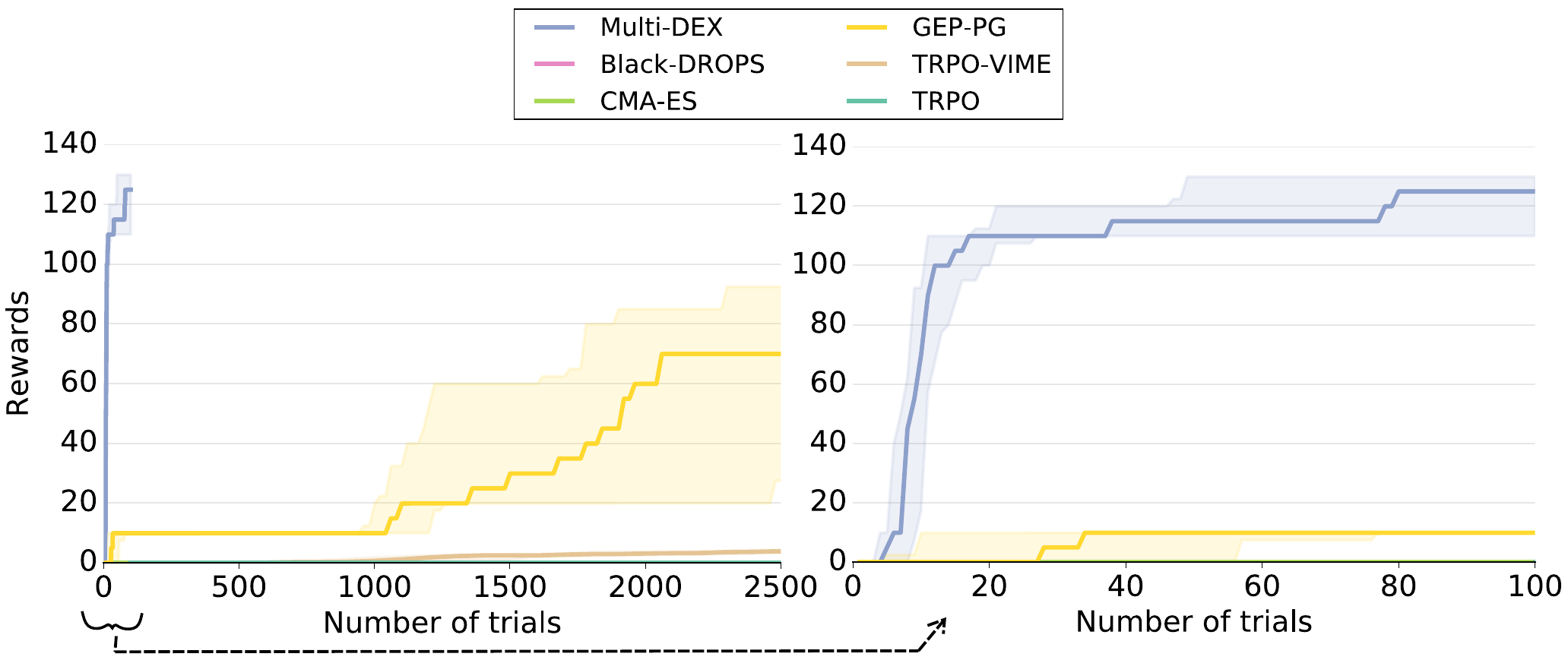}
  \caption{\label{plot:pendulum_swingup_task} Best reward found vs number of trials plot for Pendulum Swing-up Task. The plot clearly outperforms all the competing approaches and achieves very high reward (balancing the pendulum in upright position) in just 100 trials (approx 6.6 minutes of total interaction)}
\end{figure}

\begin{itemize}
  \item \textbf{State:} $\mathbf{x}_{\text{pend\_sys}} = [\theta, \dot\theta] \in \mathbb{R}^2$ , where $\theta$ is the joint angle and $\dot\theta$ is the joint angular velocity. The initial state of the system is $x_0 = [0, 0]$
  \item \textbf{Actions:} $\mathbf{u}_{\text{pend\_sys}} = [\tau] \in \mathbb{R}, -2.0 \le \tau \le 2.0$, where $\tau$ joint torque command for the arm.
  \item \textbf{Reward:} In this task, the reward is known to the algorithm as it is a function of the observable states and applied action to the system. The total reward is given by: \\
  \\
  \begin{align}
    & r(\mathbf{x}_{\text{pend\_sys}}, \tau) = r_1(\mathbf{x}_{\text{pend\_sys}}, \tau) + r_2(\mathbf{x}_{\text{pend\_sys}}, \tau)\\
    & r_1(\mathbf{x}_{\text{pend\_sys}}, \tau) =
                              \begin{cases}
                                0, &\quad\text{if } \pi - \theta < \pi/60  \\
                                +10, &\quad\text{otherwise.} \\
                              \end{cases}\\
     & r_2(\mathbf{x}_{\text{pend\_sys}},\tau) = - 0.001*\tau^2
  \end{align}
\end{itemize}

The results show that \algo{} quickly reaches to a very high positive reward with very small variance within the budget of 100 trials (Fig.~\ref{plot:pendulum_swingup_task}). On the contrary, GEP-PG and TRPO-VIME could not reach to the same quality of solutions even after 2500 trials on the system. As TRPO, Black-DROPS and CMA-ES do not have any directed exploration, they could not improve the reward and stay close to zero by minimizing the applied torque only.

\subsection{Drawer opening task with 4-DOF arm}
\begin{figure}
  \centering
  \includegraphics[width=0.7\textwidth]{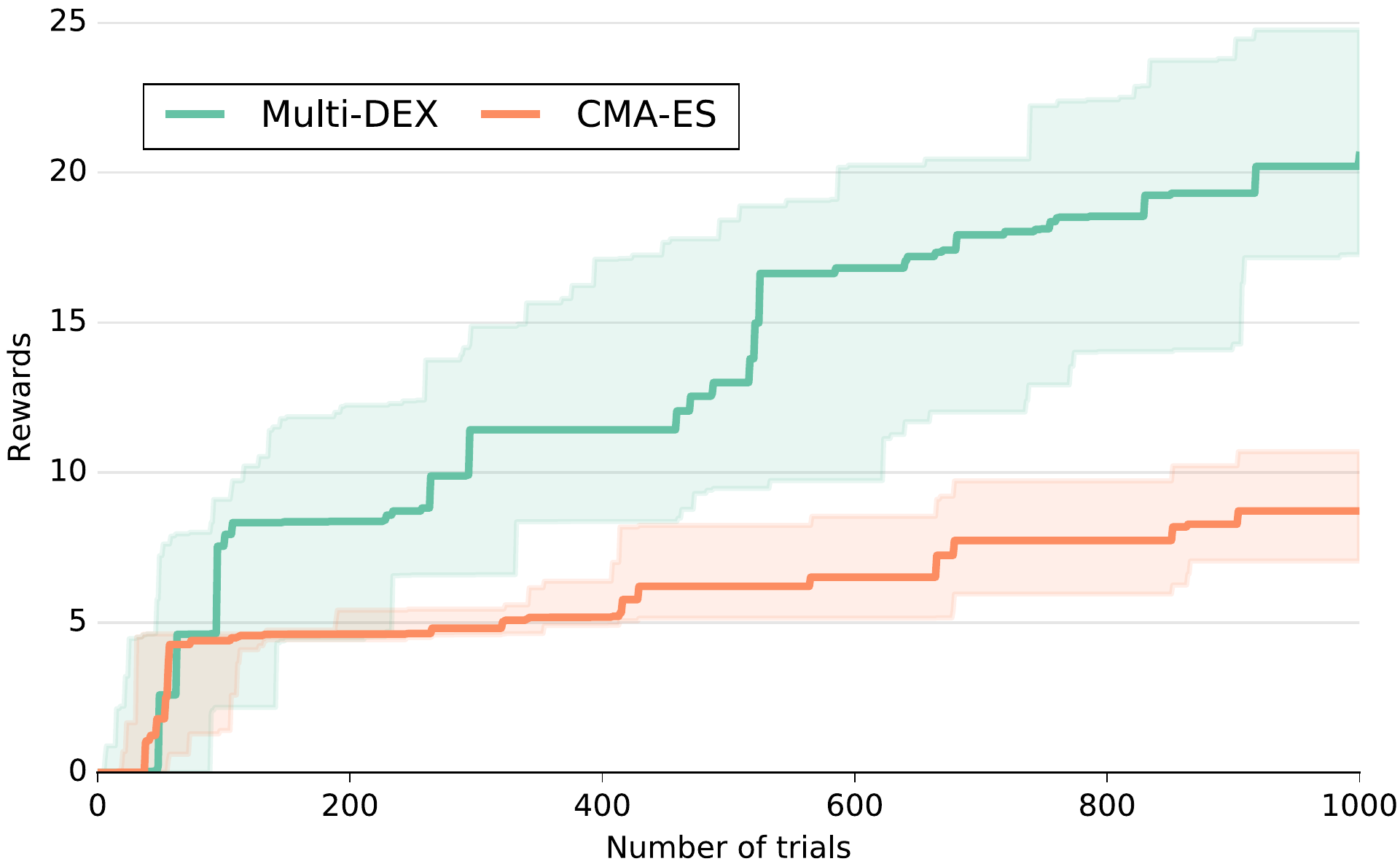}
  \caption{\label{plot:darwer_opening_task_4dof} Best reward found vs number of trials plot for drawer opening task with 4-DOF simulated robotic arm. Plot shows that both the median over the replicates with \algo{} is able solve the task in 1000 trials contrary to CMA-ES which is able to reach only half the reward achieved by \algo{})}
\end{figure}

To evaluate \algo{} on more difficult sparse reward scenarios, we ran it on the drawer opening task mentioned in the paper with a 4-dof simulated robotic arm. We compared the result with CMA-ES and found that \algo{} takes around 1000 trials on the system to solve it (median over 20 replicates). On the contrary, CMA-ES was able to reach only half the reward achieved by \algo{} in 1000 trials (Fig.~\ref{plot:darwer_opening_task_4dof}).  

\subsection{ \algo{} on non-sparse reward task}
\begin{figure}
  \centering
  \includegraphics[width=0.7\textwidth]{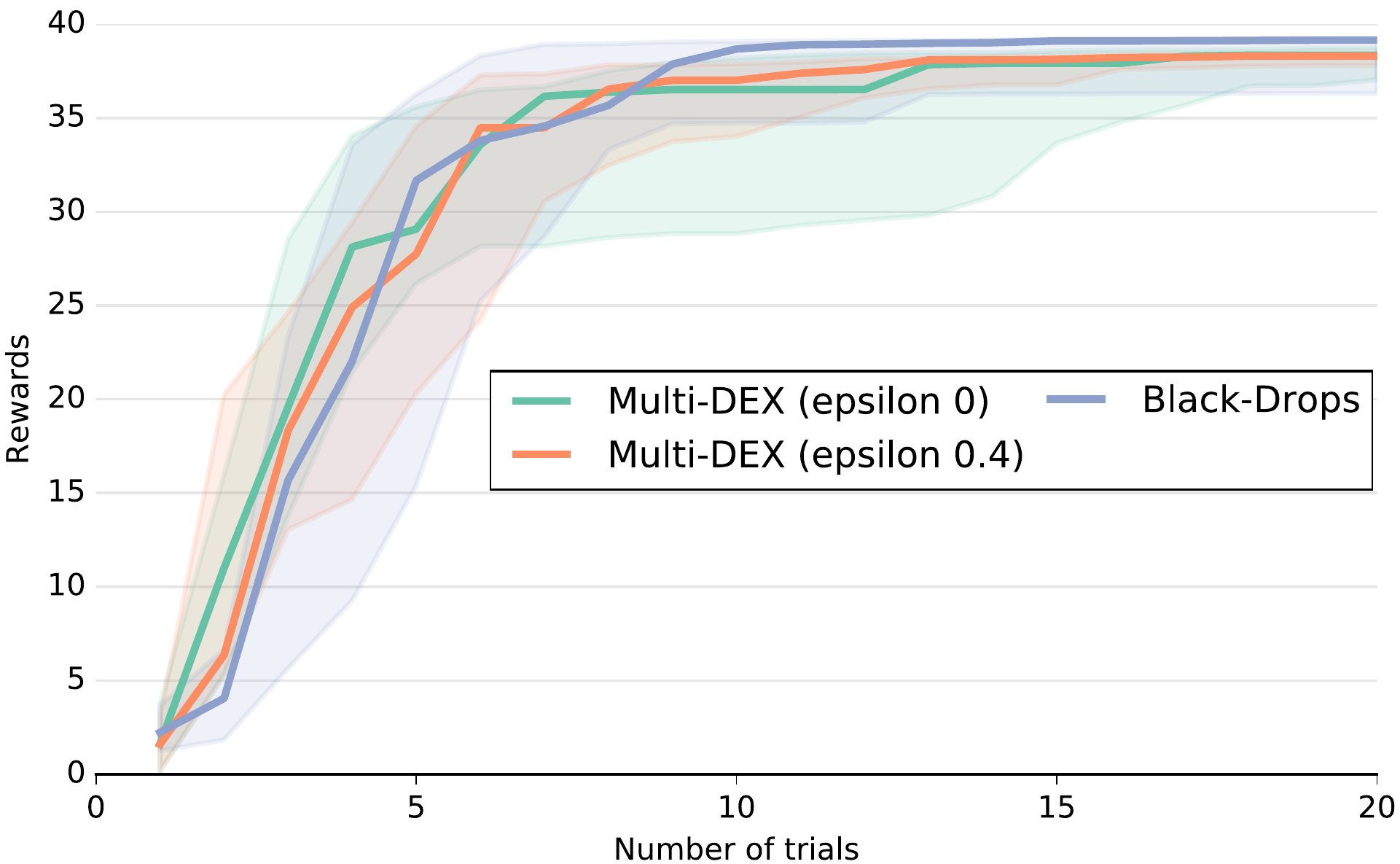}
  \caption{\label{plot:goal_reacher_5dof_task} Best reward found vs number of trials plot for single goal reaching task with a 5-DOF simulated arm. Plot shows that both the variants of \algo{} are competitive enough with Black-DROPS in this non-sparse reward scenario}
\end{figure}

We tested \algo{} on a single goal reaching task with a 5-dof simulated arm. In this task the state-space is 5 dimensional (5 joint angles of the arm) and action space is also 5 dimensional (torque inputs to the joints). The length of each episode is 4 seconds ($10$Hz control) and we used neural network policy with one hidden layer. The hidden layer had 10 neurons and thus the entire policy space is 115 dimensional. We chose a reward that is continuous in the state-space as given below:

\begin{align}
  & r(\mathbf{x}_{\text{arm}}) = \exp (-\frac{0.5}{0.2 \times 0.2} ||\mathbf{x}_{\text{goal}} - \mathbf{x}_{\text{arm}}||^2)
\end{align}

where $\mathbf{x}_{\text{goal}}$ and $\mathbf{x}_{\text{arm}}$ are the 3D coordinate vectors of the the goal and the end-effector. We compared two variants of \algo{}: with $\epsilon = 0$ and $\epsilon = 0.4$. We compared the result with standard Black-DROPS to see how efficient \algo{} is on non-sparse reward setup. This preliminary results show that both the variants of \algo{} is competitive enough with Black-DROPS (Fig.~\ref{plot:goal_reacher_5dof_task}). Point to be noted here that we plot here the median, 25 percentile and 75 percentile of "best reward so far" for the replicates. 

\putbib
\end{bibunit}

\end{document}